\documentclass[10pt,twocolumn,letterpaper]{article}

\usepackage{iccv}
\usepackage{times}
\usepackage{epsfig}
\usepackage{graphicx}
\usepackage{amsmath}
\usepackage{amssymb}
\usepackage{multirow}
\usepackage{subfigure}
\usepackage{amsfonts}
\usepackage{comment}
\usepackage{cite}
\usepackage{stfloats}
\usepackage{makecell}
\usepackage{bbding}
\usepackage{color}
\usepackage{xcolor}

\def\eg{\emph{e.g.}} 
\def\ie{\emph{i.e.}} 
 
\def\etc{\emph{etc.}}

\def\first{\textcolor{red}}
\def\second{\textcolor{blue}}
\definecolor{yellow}{RGB}{255,180,0}
\definecolor{green2}{RGB}{0,176,80}

\usepackage[pagebackref=true,breaklinks=true,colorlinks,linkcolor=red,bookmarks=false]{hyperref}

\iccvfinalcopy % *** Uncomment this line for the final submission

\def\iccvPaperID{2989} % *** Enter the ICCV Paper ID here

\ificcvfinal\fi

\begin{document}

%%%%%%%%% TITLE
\title{RINDNet: Edge Detection for Discontinuity in \\
Reflectance, Illumination, Normal and Depth}

\author{Mengyang Pu$^{1}$\thanks{The work is partially done while the author was with Stony Brook University. \quad $\dagger$ Corresponding author.}, Yaping Huang$^{1,\dagger}$, Qingji Guan$^{1}$, Haibin Ling$^{2}$\\ 
$^1$Beijing Key Laboratory of Traffic Data Analysis and Mining, Beijing Jiaotong University, China\\
$^2$Department of Computer Science, Stony Brook University, USA\\
{\tt\small \{mengyangpu, yphuang, qjguan\}@bjtu.edu.cn;\quad hling@cs.stonybrook.edu}
}

\maketitle
\ificcvfinal\thispagestyle{empty}\fi

%%%%%%%%% ABSTRACT
\begin{abstract}
   As a fundamental building block in computer vision, edges can be categorised into four types according to the discontinuity in \textit{surface-Reflectance}, \textit{Illumination}, \textit{surface-Normal} or \textit{Depth}. While great progress has been made in detecting generic or individual types of edges, it remains under-explored to comprehensively study all four edge types together. In this paper, we propose a novel neural network solution, \textit{RINDNet}, to jointly detect all four types of edges. Taking into consideration the distinct attributes of each type of edges and the relationship between them, RINDNet learns effective representations for each of them and works in three stages. In stage \uppercase\expandafter{\romannumeral1}, RINDNet uses a common backbone to extract features shared by all edges. Then in stage \uppercase\expandafter{\romannumeral2} it branches to prepare discriminative features for each edge type by the corresponding decoder. In stage \uppercase\expandafter{\romannumeral3}, an independent decision head for each type aggregates the features from previous stages to predict the initial results. Additionally, an attention module learns attention maps for all types to capture the underlying relations between them, and these maps are combined with initial results to generate the final edge detection results. For training and evaluation, we construct the first public benchmark, BSDS-RIND, with all four types of edges carefully annotated. In our experiments, RINDNet yields promising results in comparison with state-of-the-art methods. Additional analysis is presented in \href{https://github.com/MengyangPu/RINDNet}{supplementary material}.
\end{abstract}

\begin{figure}
\centering
\includegraphics[width=0.9\linewidth,height=.56\linewidth
]{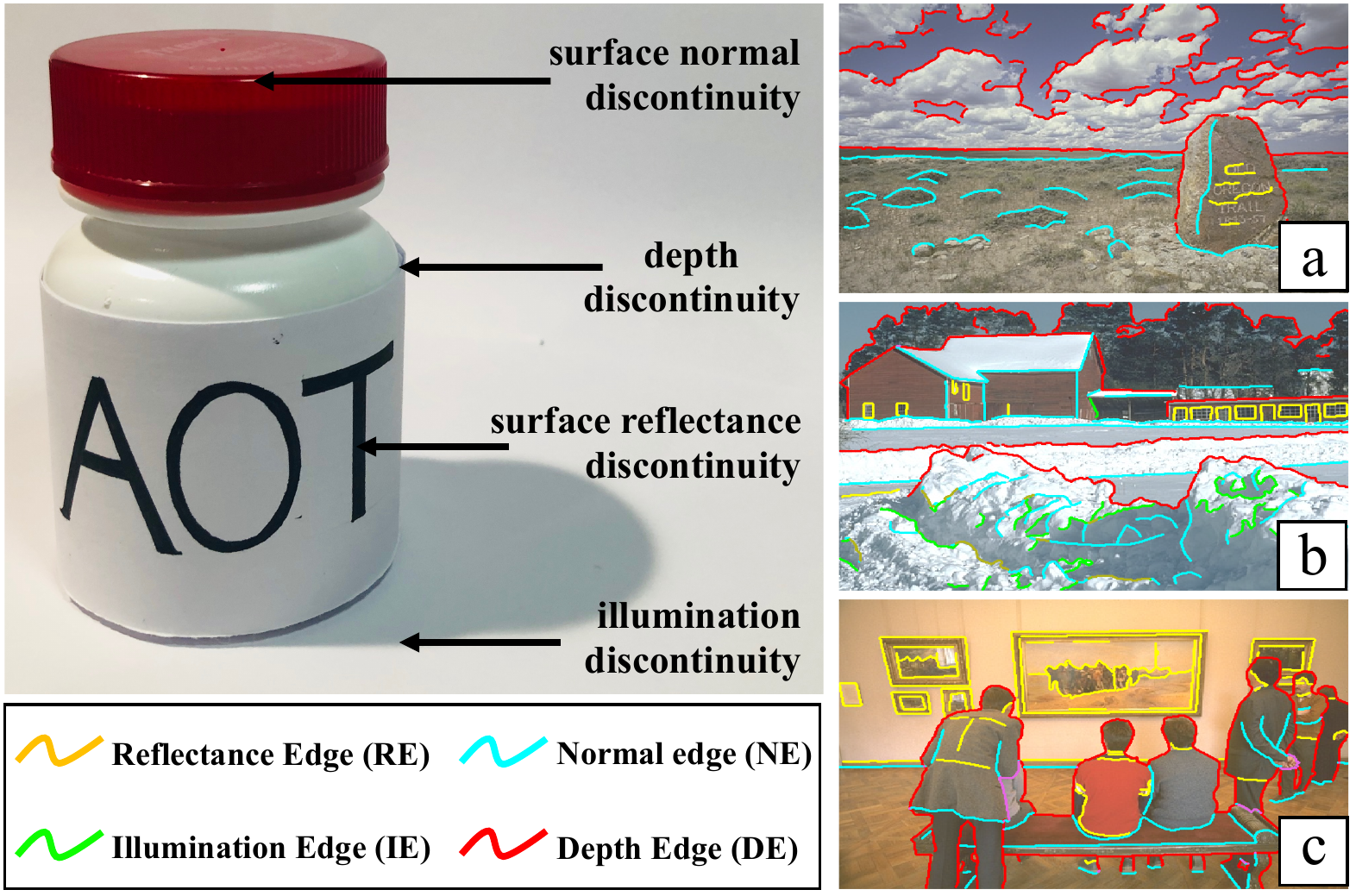}
\caption{\textbf{Left:} Edges are caused by a variety of factors \cite{marr1982vision,Seitz04wi}. \textbf{Right (a-c):} Samples of the proposed BSDS-RIND dataset.}
\label{fig:fig1}
\vspace{-13pt}
\end{figure}

%------------------------------------------------------------------------
\section{Introduction}
\label{1}

Edges play an important role in many vision tasks~\cite{YangZYPML20,wu2012strong,ramamonjisoa2020predicting,wang2015designing}. While generic edge detection~\cite{xie2015hed,liu2017rcf,he2019bdcn,wang2017ced} has been extensively studied for decades, specific-edge detection recently attracts increasing amount of efforts due to its practical applications concerning on different types of edges, such as occlusion contours~\cite{wang2016doc,wang2018doobnet,lu2019ofnet} or semantic boundaries~\cite{yu2017casenet,dff19}.

In his seminal work \cite{marr1982vision}, David Marr summarized four basic ways edges can arise: (1) \textit{surface-reflectance discontinuity}, (2) \textit{illumination discontinuity}, (3) \textit{surface-normal discontinuity}, and (4) \textit{depth discontinuity}, as shown in Fig.~\ref{fig:fig1}. Recent studies~\cite{YangZYPML20,wu2012strong,ramamonjisoa2020predicting,wang2015designing} have shown that the above types of edges are beneficial for downstream tasks. For example, pavement crack detection (reflectance discontinuity) is a critical task for intelligent transportation \cite{YangZYPML20}; shadow edge (illumination discontinuity) detection is a prerequisite for shadow removal and path detection \cite{wu2012strong}; \cite{ramamonjisoa2020predicting} and \cite{wang2015designing} show that depth edge and normal edge representation prompt refined normal and sharp depth estimation, respectively. Besides, \cite{KimTO15Joint} utilizes four types of cues simultaneously to improve the performance of depth refinement.

Despite their importance, fine-grained edges are still under-explored, especially when compared with generic edges. Generic edge detectors usually treat edges indistinguishably; while existing studies for specific edges focus on individual edge type. By contrast, the four fundamental types of edges, to our best knowledge, have never been explored in an integrated edge detection framework.

In this paper, for the first time, we propose to detect simultaneously the four types of edges, namely \textit{reflectance edge} (RE), \textit{illumination edge} (IE), \textit{normal edge} (NE) and \textit{depth edge} (DE). Although edges share similar patterns in intensity variation in images, they have different physical bases. Specifically, REs and IEs are mainly related to photometric reasons -- REs are caused by changes in material appearance (\eg, texture and color), while IEs are produced by changes in illumination (\eg, shadows, light sources and highlights). By contrast, NEs and DEs reflect the geometry changes in object surfaces or depth discontinuity. Considering the correlations and distinctions among all types of edges, we develop a CNN-based solution, named \textit{RINDNet}, for jointly detecting the above four types of edges.

RINDNet works in three stages. In stage \uppercase\expandafter{\romannumeral1}, it extracts general features and spatial cues from a backbone network for all edges. Then, in stage \uppercase\expandafter{\romannumeral2}, it proceeds with four separate decoders. Specifically, low-level features are first integrated under the guidance of high-level hints by Weight Layer (WL), and then fed into RE-Decoder and IE-Decoder to produce features for REs and IEs respectively. At the same time, NE/DE-Decoder take the high-level features as input and explore effective features. After that, these features and accurate spatial cues are forwarded to four decision heads in stage \uppercase\expandafter{\romannumeral3} to predict the initial results. Finally, the attention maps obtained by the Attention Module (AM), which captures the underlying relations between all types, are aggregated with the initial results to generate the final predictions. All these components are differentiable, making RINDNet an end-to-end architecture to jointly optimize the detection of four types of edges.

Training and evaluating edge detectors for all four types of edges request images with all such edges annotated. In this paper, we create the first known such dataset, named \textit{BSDS-RIND}, by carefully labeling images from the BSDS~\cite{arbelaez2010bsds} benchmark (see Fig.~\ref{fig:fig1}). BSDS-RIND allows the first thorough evaluation of edge detection of all four types. The proposed RINDNet shows clear advantages over previous edge detectors, both quantitatively and qualitatively. The source code, dataset, and benchmark are available at \url{https://github.com/MengyangPu/RINDNet}.

With the above efforts, our study is expected to stimulate further research along the line, and benefit more downstream applications with rich edge cues. Our contributions are summarized as follows: (1) We develop a novel end-to-end edge detector, RINDNet, to jointly detect the four types of edges. RINDNet is designed to effectively investigate shared information among different edges (\eg, through feature sharing) and meanwhile flexibly model the distinction between them (\eg, through edge-aware attention). (2) We present the first public benchmark, BSDS-RIND, dedicated to studying simultaneously the four edge types, namely reflectance edge, illumination edge, normal edge and depth edge. (3) In our experiments, the proposed RINDNet shows clear advantages over state of the arts.

%------------------------------------------------------------------------
\section{Related Works}
\label{2}

\vspace{-2mm}
\paragraph{Edge Detection Algorithms.}
Early edge detectors~\cite{kittler1983accuracy,canny1986computational,winnemoller2011xdog} obtain edges based directly on the analysis of image gradients. By contrast, learning-based methods \cite{martin2004learning,dollar2006supervised,lim2013sketch} exploit different low-level features that respond to characteristic changes, then a classifier is trained to generate edges. CNN-based edge detectors~\cite{kokkinos2015pushing,deng2020dscd,xu2017AMHNet,shen2015deepcontour,bertasius2015deepedge,bertasius2015hfl,liu2016rds,maninis2016cob,deng2018lpcb,kelm2019rcn,poma2020dexined} do not rely on hand-crafted features and achieve better performance.
Combining multi-scale and multi-level features, \cite{xie2015hed,liu2017rcf,poma2020dexined,he2019bdcn} yield outstanding progresses on generic edge detection. A novel refinement architecture is also proposed in \cite{wang2017ced} using a top-down backward refinement pathway to generate crisp edges.
Recent works~\cite{zhen2020joint,acuna2019devil,yu2018simultaneous,yang2016object} pay more attention to special types of edges. In~\cite{hariharan2011semantic} the generic object detector is combined with bottom-up contours to infer object contours. CASENet~\cite{yu2017casenet} adopts a nested architecture to address semantic edge detection. For better prediction, DFF~\cite{dff19} learns adaptive weights to generate specific features of each semantic category. For occlusion boundary detection, DOC~\cite{wang2016doc} decomposes the task into occlusion edge classification and occlusion orientation regression, then two sub-networks are used to separately perform the above two tasks. DOOBNet~\cite{wang2018doobnet} uses an encoder-decoder structure to obtain multi-scale and multi-level features, and shares the backbone features with two branches. OFNet~\cite{lu2019ofnet} considers the relevance and distinction for the occlusion edge and orientation, thus it shares the occlusion cues between two sub-networks.

\vspace{-5mm}
\paragraph{Edge Datasets.} 
Many datasets have been proposed for studying edges.
BSDS \cite{arbelaez2010bsds} is a popular edge dataset for detecting generic edges containing $500$ RGB natural images. Although each image is annotated by multiple users, they usually pay attention to salient edges related to objects. BIPED~\cite{mely2016multicue} is created to explore more comprehensive and dense edges, and contains $250$ outdoor images. NYUD~\cite{silberman2012indoor} contains $1,449$ RGB-D indoor images, and lacks edge types pertaining to outdoor scenes. Significantly, Multicue~\cite{mely2016multicue} considers the interaction between several visual cues (luminance, color, stereo, motion) during boundary detection.

Recently, SBD \cite{hariharan2011semantic} is presented for detecting semantic contours, using the images from the PASCAL VOC challenge \cite{everingham2010pascal}. Cityscapes \cite{cordts2016cityscapes} provides the object or semantic boundaries focusing on road scenes. For reasoning occlusion relationship between objects, the dataset in \cite{ren2006figure} consists of $200$ images, where boundaries are assigned with figure/ground labels. Moreover, PIOD~\cite{wang2016doc} contains $10,000$ images, each with two annotations: a binary edge map denotes edge pixels and a continuous-valued occlusion orientation map. The recent dataset in~\cite{ramamonjisoa2020predicting} annotates NYUD test set for evaluating the occlusion boundary reconstruction.

Our work is inspired by the above pioneer studies, but makes novel contributions in two aspects: the proposed RINDNet, to the best of our knowledge, is the first edge detector to jointly detect all four types of edges, and the proposed BSDS-RIND is the first benchmark with all four types of edges annotated.

%------------------------------------------------------------------------
\section{Problem Formulation and Benchmark}
\label{3}

\subsection{Problem Formulation}
\label{3.1}
Let $X \in \mathbb{R}^{3 \times W \times H}$ be an input image with ground-truth labels $\mathcal{E}=\{E^{r},E^{i},E^{n},E^{d}\}$, where $E^{r}, E^{i}, E^{n}, E^{d} \in\{0,1\}^{W \times H}$ are binary edge maps indicating the reflectance edges (REs), illumination edges (IEs), surface-normal edges (NEs) and depth edges (DEs), respectively. Our goal is to generate the final predictions $\mathcal{Y}=\{Y^{r}, Y^{i}, Y^{n}, Y^{d}\}$, where $Y^{r}, Y^{i}, Y^{n}, Y^{d}$ are the edge maps corresponding to REs, IEs, NEs and DEs, respectively. In our work, we aim to learn a CNN-based edge detector $\psi$: 
$\mathcal{Y} = \psi (X)$.

The training of $\psi$ can be done over training images by minimizing some loss functions between $\mathcal{E}$ and $\mathcal{Y}$. Therefore, a set of images with ground-truth labels are required to learn the mapping $\psi$. We contribute such annotations in this work, and the detailed processes are shown in \S\ref{3.2}.

\subsection{Benchmark}
\label{3.2}
One aim of our work is to contribute a first public benchmark, named BSDS-RIND, over BSDS images~\cite{arbelaez2010bsds}. The original images contain various complex scenes, which makes it challenging to jointly detect all four types of edges. Fig.~\ref{fig:fig1} (Right) shows some examples of our annotations.

\vspace{-3mm}
\paragraph{Edge Definitions.}
It is critical to define four types of edges for the annotation task. Above all, we give the definition of each type and illustrate it with examples.
\begin{itemize}
    \vspace{-1.6mm}\item \textbf{\textit{Reflectance Edges} (REs)} usually are caused by the changes in material appearance (\eg, texture and color) of smooth surfaces. Notably, although the edges within paintings in images (see Fig.~\ref{fig:fig1} (c)) could be classified to DEs by human visual system, these edges are assigned as REs since there is no geometric discontinuity. 
    \vspace{-1.6mm}\item \textbf{\textit{Illumination Edges} (IEs)} are produced by shadows, light sources, highlights, \etc\ (as shown in Fig.~\ref{fig:fig1}).
    \vspace{-1.6mm}\item \textbf{\textit{Normal Edges} (NEs)} mark the locations of discontinuities in surface orientation and normally arise between parts. As shown in Fig.~\ref{fig:fig1} (b), we take the edges between the building and the ground as an example, the change of depth across these two surfaces is continuous but not smooth, which is caused by the surface-normal discontinuity between them.
    \vspace{-1.6mm}\item \textbf{\textit{Depth Edges} (DEs)} are resulted by depth discontinuity, and often coincide with object silhouettes. It is difficult to measure the depth difference (\eg, Fig.~\ref{fig:fig1} (a)), thus the relative depth difference is used to determine whether an edge belongs to DEs. Although there exists the depth changes between windows and walls in the building (Fig.~\ref{fig:fig1} (b)), the small distance ratio is caused by the long distance between them and camera, so such edges are classified to REs rather than DEs.
\end{itemize}

\vspace{-6.5mm}
\paragraph{Annotation Process.}
The greatest effort for constructing a high-quality edge dataset is devoted to, not surprisingly, manual labeling, checking, and revision. For this task, we manually construct the annotations using ByLabel~\cite{qin2018bylabel}.
Two annotators collaborate to label each image. One annotator first manually labels the edges, and another annotator checks the result and may supplement missing edges. Those edges with labels are added directly to the final dataset if both annotators agree with each other. Ambiguous edges will be revised by both annotators together: the consistent annotations are given after discussion. After iterating several times, we get the final annotations. Moreover, for some edges that are difficult to determine the main factors of their formation, multiple labels are assigned for them. It is only about 53k (2\%) pixels with multi-labels in BSDS-RIND. In addition, we use the average Intersection-over-Union (IoU) score to measure agreement between two annotators, and get $0.97$, $0.92$, $0.93$ and $0.95$ for REs, IEs, NEs and DEs, respectively. The statistics show good consistency. 

\begin{figure}
\centering
\includegraphics[width=0.86\linewidth, height=.28\linewidth]{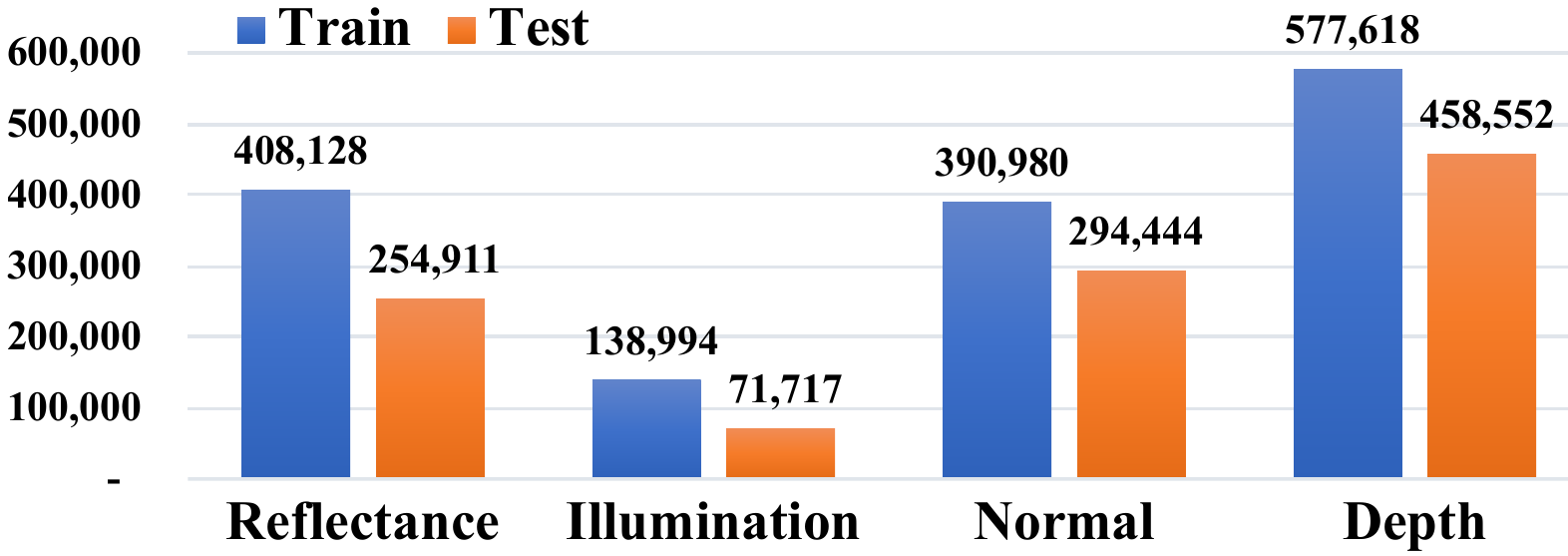}
\caption{The distribution of pixels for each type of edges on BSDS-RIND training set and testing set.}
\label{fig:edge_num}
\vspace{-13pt}
\end{figure}

With all efforts, finally, a total of $500$ images are carefully annotated, leading to a densely annotated dataset, named BSDS-RIND. Then it is split into $300$ training images, and $200$ testing images, respectively. The total number of pixels for each type on the BSDS-RIND training and testing set are reported in Fig. \ref{fig:edge_num}. Significantly, the number of edge pixels in BSDS-RIND are twice as in BSDS. Moreover, edge detection is a pixel-wise task, thus the number of samples provided by BSDS-RIND decently supports learning-based algorithms. More examples and details are given in the supplementary material.

\begin{figure*}[!t]
\centering
\includegraphics[width=0.95\linewidth, height=.44\linewidth]{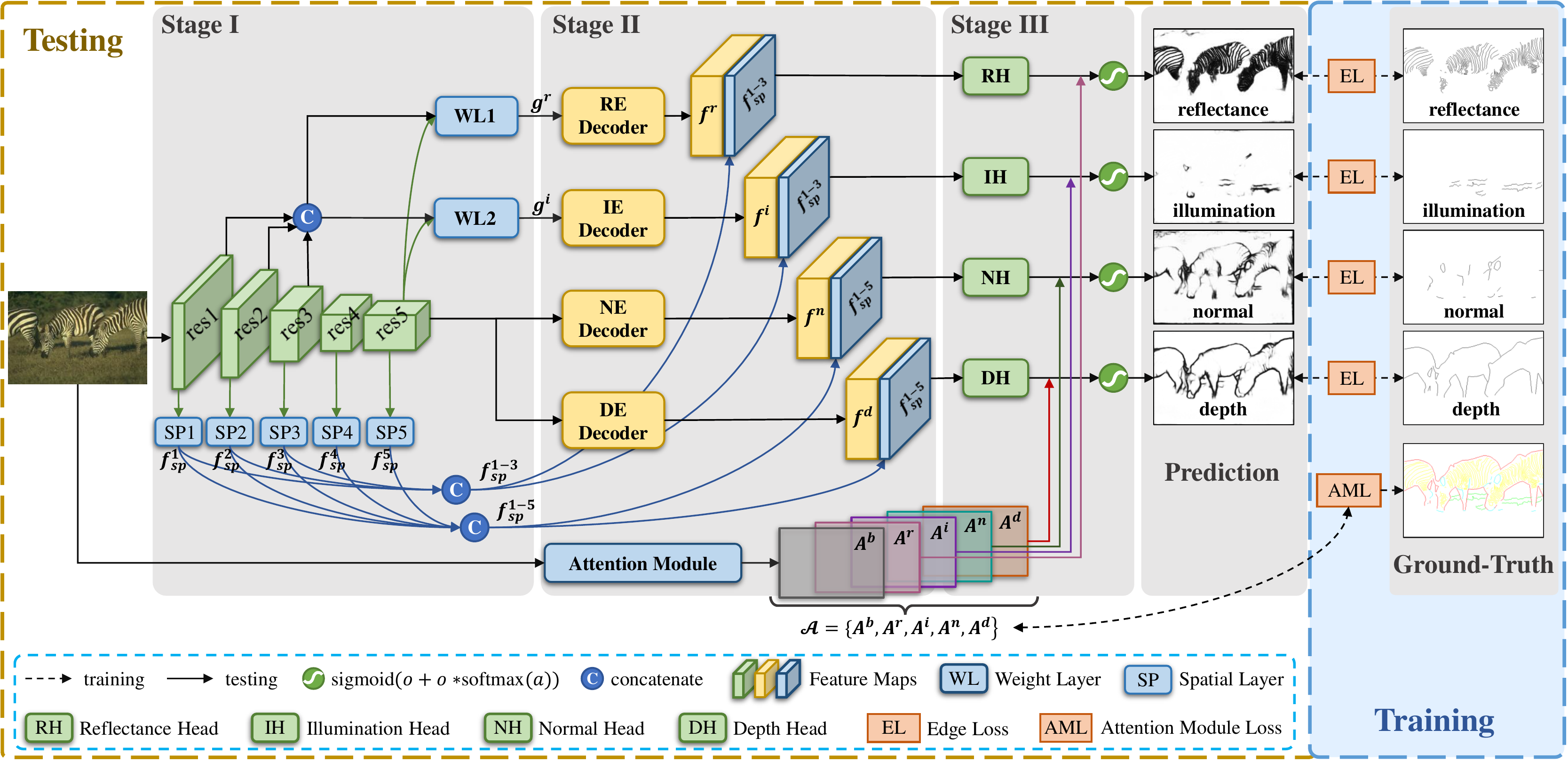}
\caption{The three-stage architecture of RINDNet.  \textbf{Stage \uppercase\expandafter{\romannumeral1}}: the input image is fed into a backbone to extract features shared with all edge types. \textbf{Stage \uppercase\expandafter{\romannumeral2}}: features across different levels are fused via Weight Layers (WLs), and are forwarded to four decoders in two clusters: RE-Decoder/IE-Decoder and NE-Decoder/DE-Decoder. \textbf{Stage \uppercase\expandafter{\romannumeral3}}: four decision heads predict the four types of initial results. In addition, the attention maps learned by the attention module are integrated into the final prediction  ($A^b$ used only in training). (Best viewed in color)}
\label{fig:fig2}
\vspace{-13pt}
\end{figure*}

%------------------------------------------------------------------------
\section{RINDNet}
\label{4}

In this work, we design an end-to-end network (\S\ref{4.1}), named \textit{RINDNet}, to learn distinctive features for optimizing the detection of four edge types jointly. Fig.~\ref{fig:fig2} shows an overview of our proposed RINDNet, which includes three stages of initial results inference (\ie, extracting common features, preparing distinctive features, and generating initial results) and final predictions integrated by an Attention Module. We also explain the loss functions and training details in \S\ref{4.2} and \S\ref{4.3}.

\subsection{Methodology}
\label{4.1}

\paragraph{Stage \uppercase\expandafter{\romannumeral1}: Extracting Common Features for All Edges.}
We first use a backbone to extract common features for all edges because these edges share similar patterns in intensity variation in images.
The backbone follows the structure of ResNet-50 \cite{he2016deepres} which is composed of five repetitive building blocks. Specifically, the feature maps from the above five blocks of ResNet-50 \cite{he2016deepres} are denoted as $res_1$, $res_2$, $res_3$, $res_4$ and $res_5$, respectively.

Then, we generate spatial cues from the above features. It is well known that different layers of CNN features encode different levels of appearance/semantic information, and contribute differently to different edge types. Specifically, bottom-layer feature maps $res_{1-3}$ focus more on low-level cues (\eg, color, texture and brightness), while top-layer maps $res_{4-5}$ are in favor of object-aware information. Thus it is beneficial to capture multi-level spatial responses from different layers of feature maps. Given multiple feature maps $res_{1-5}$, we obtain the spatial response maps:
\begin{equation}
    f_{sp}^k = \psi^{k}_{\rm sp}(res_k), \quad k \in \{1,2,3,4,5\}
\end{equation}
where the spatial responses $f_{sp}^k \in \mathbb{R}^{2 \times W \times H}$ are learned by Spatial Layer $\psi^{k}_{\rm sp}$ which is composed of one convolution layer and one deconvolution layer.

\begin{figure}
\centering
\includegraphics[width=0.9\linewidth, height=.7\linewidth]{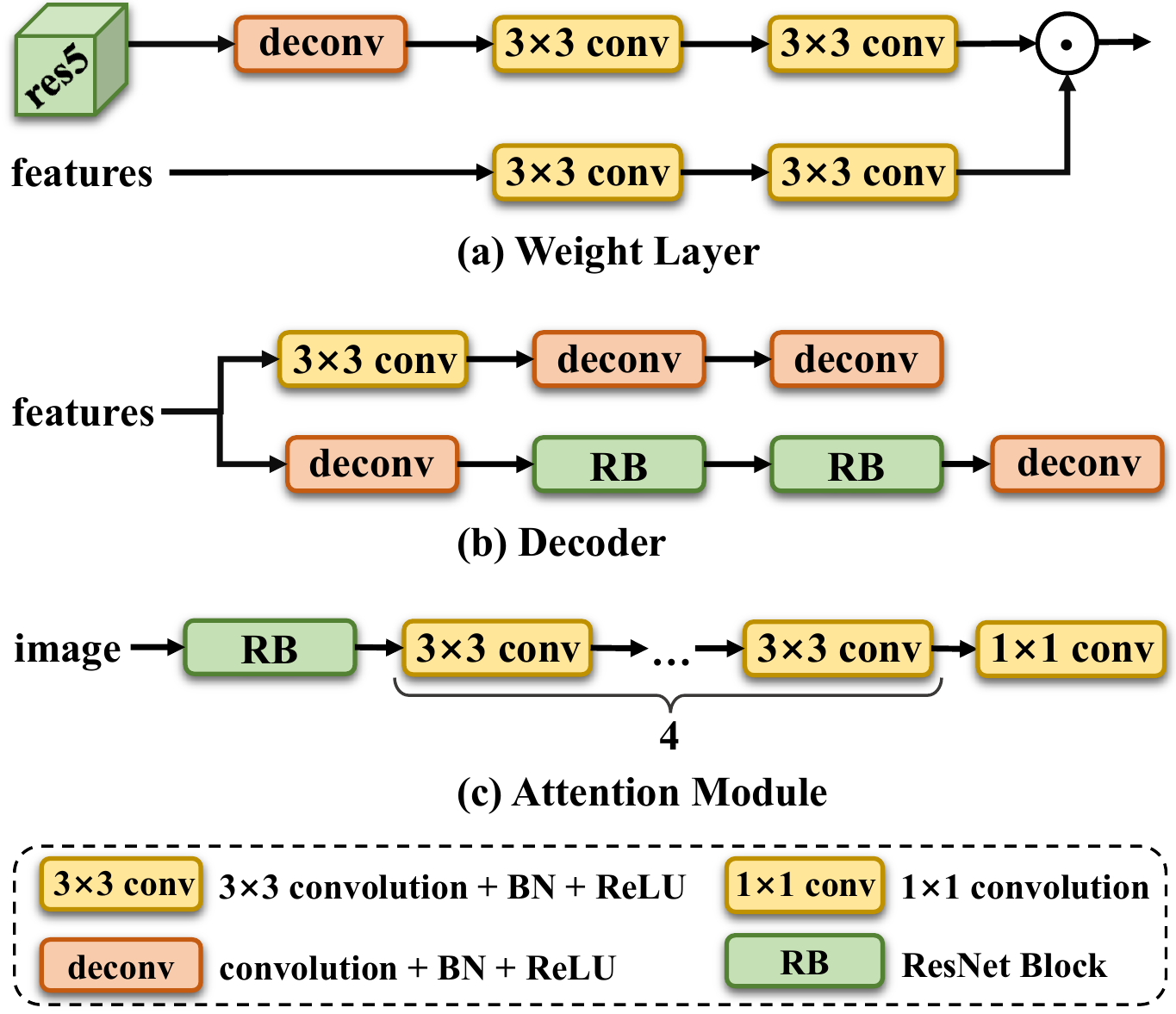}
\caption{The architectures of (a) Weight Layer, (b) Decoder and (c) Attention Module. $\odot$ is the element-wise multiplication.}
\label{fig:fig6}
\vspace{-13pt}
\end{figure}

\vspace{-3mm}
\paragraph{Stage \uppercase\expandafter{\romannumeral2}: Preparing Distinctive Features for REs/IEs and NEs/DEs.} Afterwards, RINDNet learns particular features for each edge type separately by the corresponding decoder in stage \uppercase\expandafter{\romannumeral2}. Inspired by \cite{lu2019ofnet}, we design the Decoder with two streams to recover fine location information, as shown in Fig.~\ref{fig:fig6} (b). Two-stream decoder can work collaboratively and learn more powerful features from different views in the proposed architecture. Although four decoders have the same structure, some special designs are proposed for different types of edges, and we will give the detailed descriptions below. To distinguish each type of edges reasonably and better depict our work, we next cluster the four edge types into two groups, \ie, REs/IEs and NEs/DEs, to prepare features for them respectively.

\textbf{\textit{REs and IEs.}} In practice, the low-level features (\eg, $res_{1-3}$) capture detailed intensity changes that are often reflected in REs and IEs. Besides, REs and IEs are related to the global context and surrounding objects provided by the high-level features (\eg, $res_{5}$). Thus, it is desirable that semantic hints may give the felicitous guidance to aware the intensity changes, before forwarding to the Decoder. Moreover, it is notable that simply concatenating the low-level and high-level features may be too computationally expensive due to the increased number of parameters. We therefore propose the Weight Layer (WL) to adaptively fuse the low-level features and high-level hints in a learnable manner, without increasing the dimension of the features. 

As shown in Fig.~\ref{fig:fig6} (a), WL contains two paths: the first path receives high-level feature $res_{5}$ to recover high resolution through a deconvolution layer, and then two $3\times3$ convolution layers with Batch Normalization (BN) and ReLU excavate adaptive semantic hints; another path is implemented as two convolution layers with BN and ReLU, which encodes low-level features $res_{1-3}$. Afterwards, they are fused by element-wise multiplication. Formally, given the low-level features $res_{1-3}$ and high-level hints $res_{5}$, we generate the fusion features for REs and IEs individually,
\begin{equation}
\begin{array}{ll}
    g^{r}= \psi^{r}_{\rm wl} \big(res_5, [res_1,res_2,\rm up (res_3)]\big), \\
    g^{i}= \psi^{i}_{\rm wl} \big(res_5, [res_1,res_2,\rm up (res_3)]\big),
\end{array}
\end{equation}
where the WL of REs and IEs are indicated as $\psi^{r}_{\rm wl}$ and $\psi^{i}_{\rm wl}$ respectively, $g^{r}$/$g^{i}$ are the fusion features for REs/IEs, and $[\cdot]$ is the concatenation. Note that, the resolution of $res_{3}$ is smaller than $res_{1}$ and $res_{2}$, so one up-sampling operation $\rm up(\cdot)$ is used on $res_{3}$ to increase resolution before feature concatenation.
Next, the fusion features are fed into the corresponding Decoder to generate specific features with accurate location information separately for IEs and REs,
\begin{equation}
\begin{array}{ll}
    f^{r} = \psi^{r}_{\rm deco}(g^{r}) , &
    f^{i} = \psi^{i}_{\rm deco}(g^{i}) , \\
\end{array}
\end{equation}
where $\psi^{r}_{\rm deco}$ and $\psi^{i}_{\rm deco}$ indicate Decode of REs and IEs respectively, and $f^{r}$/$f^{i}$ are decoded feature maps for REs/IEs.

\textbf{\textit{NEs and DEs.}} Since the high-level features (\eg, $res_5$) express strong semantic responses that are usually epitomized in NEs and DEs, we utilize $res_5$ to obtain the particular features for NEs and DEs,
\begin{equation}
\begin{array}{ll}
    f^{n} = \psi^{n}_{\rm deco}(res_5) , &
    f^{d} = \psi^{d}_{\rm deco}(res_5) ,\\
\end{array}
\end{equation}
where NE-Decode and DE-Decoder are denoted as $\psi^{n}_{\rm deco}$ and $\psi^{d}_{\rm deco}$ respectively, and $f^{n}$/$f^{d}$ are the decoded features of NEs/DEs. Since DEs and NEs commonly share some relevant geometry cues, we share the weights of the second stream of NE-Decoder and DE-Decoder to learn the collaborative geometry cues. At the same time, the first stream of NE-Decoder and DE-Decoder is responsible for learning particular features for REs and DEs, respectively.

\vspace{-3mm}
\paragraph{Stage \uppercase\expandafter{\romannumeral3}: Generating Initial Results.}
We predict the initial results for each type of edges by the respective decision head in final stage. The features from previous stages, containing rich location information of edges, can be used to predict edges. Specifically, we concatenate the decoded features $f^{r}$/$f^{i}$ with spatial cues $f_{sp}^{1-3}$ to predict REs/IEs,
\begin{equation}
\begin{array}{ll}
    O^{r} = \psi^{r}_{\rm h}\big([f^{r}, f_{sp}^{1-3}]\big) , & 
    O^{i} = \psi^{i}_{\rm h}\big([f^{i}, f_{sp}^{1-3}]\big) , 
\end{array}
\end{equation}
where $O^{r}$/$O^{i}$ are the initial predictions of REs/IEs. The decision heads of REs and IEs, named $\psi^{r}_{\rm h}$ and $\psi^{i}_{\rm h}$ respectively, are modeled as a $3\times3$ convolution layer and a $1\times1$ convolution layer. Note that REs and IEs do not directly rely on the location cues provided by top-layer, thus spatial cues $f_{sp}^{4-5}$ are not used for them. By contrast, all spatial cues $f_{sp}^{1-5}$ are concatenated with the decoded features to generate initial results for NEs and DEs, respectively,
\begin{equation}
\begin{array}{ll}
    O^{n} = \psi^{n}_{\rm h}\big([f^{n}, f_{sp}^{1-5}]\big) , &
    O^{d} = \psi^{d}_{\rm h}\big([f^{d}, f_{sp}^{1-5}]\big) , 
\end{array}
\end{equation}
where $\psi^{n}_{\rm h}$ and $\psi^{d}_{\rm h}$ respectively indicate the decision heads of NEs and DEs, which are composed of three $1 \times 1$ convolutional layers to integrate hints at each position. In summary, $\mathcal{O}=\{O^{r},O^{i},O^{n},O^{d}\}$ denotes the initial result set.

\begin{table*}
\caption{Quantitative comparison for REs, IEs, NEs, DEs and Average (best viewed in color: ``\first{\bf{red}}'' for best, and ``\second{\bf{blue}}'' for second best).}
  \centering
  \small
  \renewcommand\tabcolsep{3.5pt}
  \renewcommand\arraystretch{0.9}
  \begin{tabular}{|l|ccc|ccc|ccc|ccc|ccc|}
    \hline
    \multirow{2}{*}{Method}
    &\multicolumn{3}{c|}{Reflectance} & \multicolumn{3}{c|}{Illumination} & \multicolumn{3}{c|}{Normal} &\multicolumn{3}{c|}{Depth} &\multicolumn{3}{c|}{Average}\\  
    \cline{2-16}
                    & ODS   & OIS   & AP       & ODS   & OIS   & AP        & ODS   & OIS   & AP        & ODS   & OIS   & AP       & ODS   & OIS   & AP\\
    \hline
    HED~\cite{xie2015hed}             & 0.412 & 0.466 & 0.343    & 0.256 & 0.290 & 0.167     & 0.457 & 0.505 & \bf{\second{0.395}}     & 0.644 & 0.679  & 0.667    & 0.442 & 0.485 & \bf{\second{0.393}}\\
    CED~\cite{wang2017ced}                      & 0.429 & 0.473 & 0.361	 & 0.228 & 0.286 & 0.118     & 0.463 & 0.501 & 0.372    & 0.626	& 0.655	& 0.620      & 0.437 & 0.479 & 0.368\\
    RCF~\cite{liu2017rcf}             & 0.429 & 0.448 & 0.351    & 0.257 & 0.283 & \bf{\first{0.173}}     & 0.444 & 0.503 & 0.362     & 0.648 & 0.679  & 0.659    & 0.445 & 0.478 & 0.386\\
    BDCN~\cite{he2019bdcn}            & 0.358 & 0.458 & 0.252    & 0.151 & 0.219 & 0.078     & 0.427 & 0.484 & 0.334     & 0.628 & 0.661  & 0.581    & 0.391 & 0.456 & 0.311\\
    DexiNed~\cite{poma2020dexined}                   & 0.402 & 0.454 & 0.315	 & 0.157 & 0.199 & 0.082     & 0.444 & 0.486 & 0.364     & 0.637 & 0.673  & 0.645   & 0.410 & 0.453 & 0.352\\
    CASENet~\cite{yu2017casenet}      & 0.384 & 0.439 & 0.275    & 0.230 & 0.273 & 0.119     & 0.434 & 0.477 & 0.327     & 0.621 & 0.651  & 0.574    & 0.417 & 0.460 & 0.324\\
    DFF~\cite{dff19}                  & \bf{\second{0.447}} & 0.495 & 0.324    & \bf{\first{0.290}} & \bf{\first{0.337}} & 0.151     & \bf{\second{0.479}} & \bf{\second{0.512}} & 0.352     & \bf{\second{0.674}} & \bf{\second{0.699}}  & 0.626    & \bf{\second{0.473}} & \bf{\second{0.511}} & 0.363\\
    *DeepLabV3+~\cite{chen2018deeplabv3}    & 0.297 & 0.338 & 0.165    & 0.103 & 0.150 & 0.049     & 0.366 & 0.398 & 0.232     & 0.535 & 0.579  & 0.449    & 0.325 & 0.366 & 0.224\\
    *DOOBNet~\cite{wang2018doobnet}         & 0.431 & 0.489 & 0.370    & 0.143 & 0.210 & 0.069     & 0.442 & 0.490 & 0.339     & 0.658 & 0.689  & 0.662    & 0.419 & 0.470 & 0.360\\
    *OFNet~\cite{lu2019ofnet}         & 0.446 & 0.483 & \bf{\second{0.375}}    & 0.147 & 0.207 & 0.071     & 0.439 & 0.478 & 0.325     & 0.656 & 0.683  & \bf{\second{0.668}}    & 0.422 & 0.463 & 0.360\\
    \hline
    DeepLabV3+~\cite{chen2018deeplabv3}     & 0.444 & 0.487 & 0.356    & 0.241 & \bf{\second{0.291}} & 0.148     & 0.456 & 0.495 & 0.368     & 0.644 &0.671  & 0.617    & 0.446 & 0.486 & 0.372\\
    DOOBNet~\cite{wang2018doobnet}    & 0.446 & \bf{\second{0.503}} & 0.355   & 0.228 & 0.272 & 0.132     & 0.465 & 0.499 & 0.373     & 0.661 & 0.691 & 0.643     & 0.450 & 0.491 & 0.376\\
    OFNet~\cite{lu2019ofnet}          & 0.437 & 0.483 & 0.351   & 0.247 & 0.277 & 0.150     & 0.468 & 0.498 & 0.382     & 0.661 & 0.687 & 0.637     & 0.453 & 0.486 & 0.380\\
    
    \hline
    RINDNet (Ours) & \bf{\first{0.478}} & \bf{\first{0.521}} & \bf{\first{0.414}}    &\bf{\second{0.280}} &\bf{\first{0.337}} &\bf{\second{0.168}}      &\bf{\first{0.489}} &\bf{\first{0.522}} &\bf{\first{0.440}}     &\bf{\first{0.697}} &\bf{\first{0.724}} &\bf{\first{0.705}}      &\bf{\first{0.486}} &\bf{\first{0.526}} &\bf{\first{0.432}}\\
    \hline
  \end{tabular}
  \label{tab:tab1}
\vspace{-8pt}
\end{table*}

\begin{figure*}[!t]
\centering
\includegraphics[width=0.99\linewidth, height=.20\linewidth
]{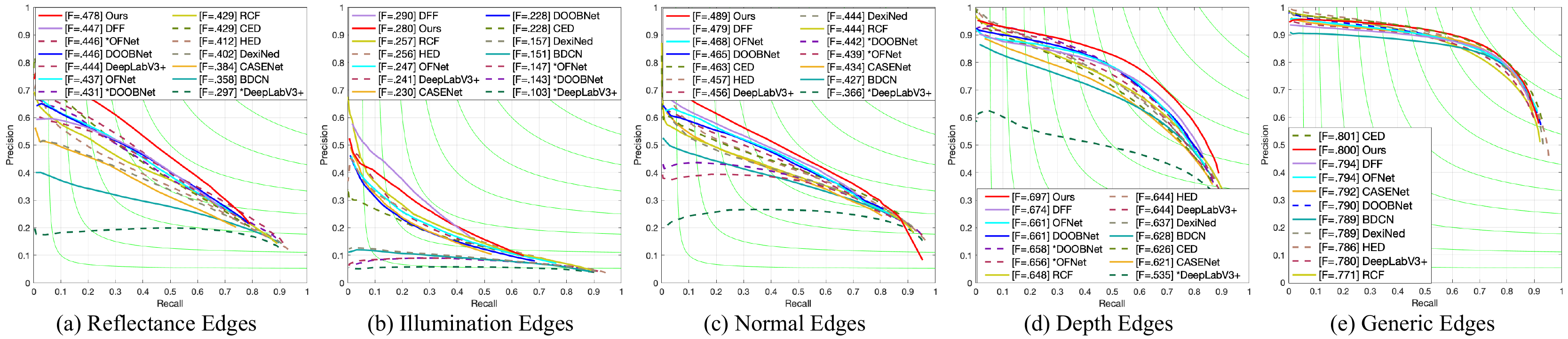}
\caption{Evaluation results on BSDS-RIND for (a) REs, (b) IEs, (c) NEs, (d) DEs and (e) generic edges.}
\label{fig:fig3}
\vspace{-8pt}
\end{figure*}

\vspace{-3mm}
\paragraph{Attention Module.} Finally, RINDNet integrates initial results with attention maps obtained by Attention Module (AM) to generate the final results. Since different types of edges are reflected in different locations, when predicting each type of edges, it is necessary to pay more attention to the related locations. Fortunately, the edge annotations provide the label of each location. Accordingly, the proposed AM could infer the spatial relationships between multiple labels with pixel-wise supervision by the attention mechanism. The attention maps could be used to activate the responses of related locations. Formally, given the input image $X$, AM learns spatial attention maps,
\begin{equation} \label{eq:eq4.4}
    \mathcal{A} =\{A^{b},A^{r},A^{i},A^{n},A^{d}\} ={\rm softmax} \big(\psi_{\rm att}(X) \big),\\
\end{equation}
where $\mathcal{A}$ is the normalized attention maps by a softmax function, and $A^{b},A^{r},A^{i},A^{n},A^{d} \in [0, 1]^{W \times H}$ are the attention maps corresponding to background, REs, IEs, NEs and DEs respectively. Obviously, if a label is tagged to one pixel, the location of this pixel should be assigned with higher attention values. The AM $\psi_{\rm att}$ is achieved by the first building block of ResNet, four $3 \times 3$ convolution layers (each layer is followed by ReLU and BN operations), and one $1 \times 1$ convolution layers, as shown in Fig. \ref{fig:fig6} (c). 

Finally, the initial results are integrated with the attention maps to generate the final results $\mathcal{Y}$,
\begin{equation} 
    \mathcal{Y} = {\rm sigmoid}\big(\mathcal{O} \odot (1 + A^{\{r,i,n,d\}})\big),
\end{equation}
where $\odot$ is the element-wise multiplication.

\subsection{Loss Function}
\label{4.2}
\paragraph{Edge Loss.}
We use the loss function presented in \cite{wang2018doobnet} to supervise the training of our edge predictions:
\begin{equation}
\mathcal{L}_{\rm e}(\mathcal{Y},\mathcal{E})= \sum_{k \in {\left\{r,i,n,d\right\}}} \ell_{\rm e} \left (Y^{k},E^{k} \right),
\end{equation}
\begin{equation}\label{eq10}
\begin{split}\ell_{\rm e} \left (Y,E \right)  & = - \sum_{i,j} \Big(   E_{i,j}\alpha_{1}\beta^{(1-Y_{i,j})^{\gamma_{1}}}{\rm log}(Y_{i,j}) \\
    &  + (1-E_{i,j})(1-\alpha_{1})\beta^{Y^{\gamma_{1}}}{\rm log}(1-Y_{i,j}) \Big), 
\end{split}
\end{equation}
where $\mathcal{Y}=\{Y^{r}, Y^{i}, Y^{n}, Y^{d}\}$ is the final prediction, $\mathcal{E} = \{E^{r},E^{i},E^{n},E^{d}\}$ is the corresponding ground-truth label, and $E_{i,j}$/$Y_{i,j}$ are the $(i,j)^{th}$ element of matrix $E$/$Y$ respectively.
Moreover, $\alpha_{1}=|E_{-}|/|E|$ and $1-\alpha_{1}=|E_{+}|/|E|$, where $E_{-}$ and $E_{+}$ denote the non-edge and edge ground truth label sets, respectively. In addition, $\gamma_{1}$ and $\beta$ are the hyperparameters. We drop the superscript $k$ in Eq. \ref{eq10} and Eq. \ref{eq13} for simplicity.

\vspace{-5mm}
\paragraph{Attention Module Loss.}
Since the pixel-wise edge annotations provide spatial labels, it is easy to obtain the ground truth of attention.
Let $\mathcal{T} = \{T^{b},T^{r},T^{i},T^{n},T^{d}\}$ be the ground-truth label of attention, where $T^{b}$ specifies the non-edge pixels. $T^{b}_{i,j} = 1$ if the $(i,j)^{th}$ pixel is located on non-edge/background, otherwise $T^{b}_{i,j} = 0$.
$T^{r},T^{i},T^{n},T^{d}$ indicate attention labels of REs, IEs, NEs and DEs respectively, which are obtained from $\mathcal{E} = \{E^{r},E^{i},E^{n},E^{d}\}$,
\begin{equation}
\label{eq:att_gt}
T_{i,j}^k = \left\{\begin{array}{ll}
    E_{i,j}^k, & {\rm if} \sum_{k} E_{i,j}^k =1, k\in\{r,i,n,d\} \\
    255, & {\rm if} \sum_{k} E_{i,j}^k > 1, k\in\{r,i,n,d\} \\
  \end{array} \right.,
\end{equation}
where $k$ denotes the type of edges, $T_{i,j}^k$ and $E_{i,j}^k$ indicate the attention label and edge label of the $(i,j)^{th}$ pixel, respectively. The attention label equals the edge label if one pixel is only assigned one type of edge label, or it will be tagged 255 that will be ignored during training if one pixel has multiple types. It is should be noted that multi-labeled edges are used when training four decision heads for each type of edges, and only excluded when training the AM. The loss function $\mathcal{L}_{\rm att}$ of the AM is formulated as:
\begin{equation}
\mathcal{L}_{\rm att}(\mathcal{A},\mathcal{T}) = 
\sum_{k \in {\left\{b,r,i,n,d\right\}}} \ell_{\rm foc} \left (A^{k},T^{k} \right),
\end{equation}
\begin{equation}\label{eq13}
\begin{split}
\ell_{\rm foc} &\left (A,T \right)  = - \sum_{i,j} \big( T_{i,j}\alpha_{2}(1-A_{i,j})^{\gamma_{2}} {\rm log}(A_{i,j}) \\
& + (1-T_{i,j})(1-\alpha_{2})A_{i,j}^{\gamma_{2}} {\rm log}(1-A_{i,j})\big),
\end{split}
\end{equation}
where $\ell_{\rm foc}$ indicates the Focal Loss \cite{lin2017focal} and $\mathcal{A}$ is the output of Attention Module. Note that $\alpha_{2}$ and $\gamma_{2}$ are a balancing weight and a focusing parameter, respectively.

\textbf{Total Loss.} Finally, we optimize RINDNet by minimizing the total loss defined as:
\begin{equation}
\label{eq:eq4}
\mathcal{L} = \lambda \mathcal{L}_{\rm e} + (1-\lambda) \mathcal{L}_{\rm att} ,
\end{equation}
where $\lambda$ is the weight for balancing the two losses.

\begin{table*}
  \caption{Ablation study to verify the effectiveness of each component in our proposed RINDNet.}
  \label{tab:wlam}
  \centering
  \small
  \renewcommand\tabcolsep{3.7pt}
  \renewcommand\arraystretch{0.9}
  \begin{tabular}{|l|ccc|ccc|ccc|ccc|ccc|}
    \hline
    \multirow{2}{*}{Method}
    &\multicolumn{3}{c|}{Reflectance} & \multicolumn{3}{c|}{Illumination} & \multicolumn{3}{c|}{Normal} &\multicolumn{3}{c|}{Depth} &\multicolumn{3}{c|}{Average}\\  
    \cline{2-16}
                        & ODS   & OIS   & AP       & ODS   & OIS   & AP         & ODS   & OIS   & AP        & ODS   & OIS   & AP    & ODS   & OIS   & AP\\
    \hline
    Ours             & 0.478 & 0.521 & 0.414     & 0.280 & 0.337 & 0.168     & 0.489 & 0.522 & 0.440     & 0.697 & 0.724 & 0.705     & 0.486 & 0.526 & 0.432\\
    Ours w/o WL    & 0.422 & 0.468 & 0.357    & 0.280 & 0.321 & 0.180      & 0.476 & 0.515 & 0.425     & 0.693 & 0.713 & 0.700 & 0.468 & 0.504 & 0.416\\
    Ours w/o AM   & 0.443 & 0.494 & 0.338    & 0.268 & 0.327 & 0.139      & 0.473 & 0.506 & 0.378     & 0.670 & 0.699 & 0.649 & 0.464 & 0.507 & 0.376\\   
    Ours w/o AM\&WL & 0.409 & 0.460 & 0.316  & 0.277 & 0.331 & 0.178      & 0.471 & 0.507 & 0.389     & 0.677 & 0.707 & 0.662 & 0.459 & 0.501 & 0.386\\
    \hline
  \end{tabular}
  \vspace{-13pt}
\end{table*}

\subsection{Training Details}
\label{4.3}
Our network is implemented using PyTorch \cite{paszke2017automatic} and finetuned from a ResNet-50 model pre-trained on ImageNet \cite{DengDSLL009}. Specifically, we adopt the Stochastic Gradient Descent optimizer with momentum=$0.9$, initial learning rate=$10^{-5}$, and we decay it by the ``poly'' policy on every epoch. We train the model for $70$ epochs on one GPU with a batch size of $4$. Moreover, we set $\beta=4$ and $\gamma_{1}=0.5$ for $\mathcal{L}_{\rm e}$; $\alpha_{2}=0.5$ and $\gamma_{2}=2$ for $\mathcal{L}_{\rm att}$; and $\lambda=0.1$ for total loss. In addition, following \cite{wang2018doobnet}, we augment our dataset by rotating each image by four different angles of $\{0,90,180,270\}$ degrees. Each image is randomly cropped to $320\times320$ during training while retaining the original size during testing.

%------------------------------------------------------------------------
\section{Experiment Evaluation}
\label{5}
We compare our model with $10$ state-of-the-art edge detectors. HED \cite{xie2015hed}, RCF \cite{liu2017rcf}, CED \cite{wang2017ced}, DexiNed \cite{poma2020dexined}, and BDCN \cite{he2019bdcn} exhibit excellent performance in general edge detection; DeepLabV3+ \cite{chen2018deeplabv3}, CASENet \cite{yu2017casenet} and DFF \cite{dff19} show outstanding accuracy on semantic edge detection; DOOBNet \cite{wang2018doobnet} and OFNet \cite{lu2019ofnet} yield competitive results for occlusion edge detection. 
All models are trained on $300$ training images and evaluated on $200$ test images. In addition, more qualitative results are provided in the supplementary material. 
We evaluate these models with three metrics introduced by \cite{arbelaez2010bsds}: fixed contour threshold (ODS), per-image best threshold (OIS), and average precision (AP). Moreover, a non-maximum suppression \cite{canny1986computational} is performed on the predicted edge maps before evaluation.

\subsection{Experiments on Four Types of Edges}
\vspace{-2mm}

\paragraph{Comparison with State of the Arts.} To adapt existing detectors for four edge types simultaneously, they are modified in two ways: (1) The output $Y \!\in\! \{0,1\}^{W \times H}$ is changed to $\mathcal{Y} \!\in\! \{0,1\}^{4 \times W \times H}$. In particular, for DeepLabV3+ focusing on segmentation, the output layer of DeepLabV3+ is replaced by an edge path (same as DOOBNet~\cite{wang2018doobnet} and OFNet~\cite{lu2019ofnet}, containing a sequence of four $3\!\times\!3$ convolution blocks and one $1\!\times\!1$ convolution layer) to predict edge maps. As shown in Table \ref{tab:tab1}, ten compared models are symbolized as HED, RCF, CED, DexiNed, BDCN, CASENet, DFF, *DeepLabV3+, *DOOBNet and *OFNet, respectively. (2) DeepLabV3+, DOOBNet and OFNet only provide one edge prediction branch without structure suitable for multi-class predictions, thus we provide the second modification: the last edge prediction branch is expanded to four, and each branch predicts one type of edges. The modification is similar to the prediction approach of our model and aims to explore the capabilities of these models. They are symbolized as DeepLabV3+, DOOBNet, and OFNet, respectively.

Table~\ref{tab:tab1} and Fig.~\ref{fig:fig3} present the F-measure of the four types of edges and their averages. We observe that the proposed RINDNet outperforms other detectors over most metrics across the dataset. Essentially, \cite{xie2015hed,liu2017rcf,wang2017ced,he2019bdcn,poma2020dexined,wang2018doobnet,lu2019ofnet} are designed for generic edge detection, so the specific features of different edges are not fully explored. Even if we extend the prediction branch of OFNet~\cite{lu2019ofnet}, DOOBNet~\cite{wang2018doobnet} and DeepLabV3+~\cite{chen2018deeplabv3} to four for learning specific features respectively, the results are still unsatisfactory. DFF~\cite{dff19} could learn the specific cues in a certain extent by introducing the dynamic feature fusion strategy, but the performance is still limited. On the contrary, our proposed RINDNet achieves promising results by extracting the corresponding distinctive features based on different edge attributes.

\begin{table}
\caption{Ablation study on the choices of features or spatial cues from different layers for the proposed RINDNet.}
\centering
\small
\renewcommand\tabcolsep{4.9pt}
\renewcommand\arraystretch{1.0}
\begin{tabular}{|c|c|c|ccc|}
    \hline
    \multirow{2}{*}{Reference} & \multirow{2}{*}{RE\&IE} & \multirow{2}{*}{NE\&DE} &\multicolumn{3}{c|}{Average} \\
    \cline{4-6}
            &        &                 & ODS   & OIS   & AP\\
    \hline
    \hline
    \multirow{4}{*}{\makecell[c]{ Different-Layer \\ Features \\}} 
    & $res_{1-3}$  & $res_{5}$  & 0.486 & 0.526 & 0.432\\
    & $res_{1-3}$  & $res_{1-3}$  & 0.467 & 0.499 & 0.422\\
    & $res_{5}$  & $res_{1-3}$  & 0.452 & 0.482 & 0.381\\   
    & $res_{5}$  & $res_{5}$  & 0.464 & 0.489 & 0.396\\
    \hline
    \hline
    \multirow{4}{*}{Spatial Cues} 
    & $f_{sp}^{1-3}$  & $f_{sp}^{1-5}$  & 0.486 & 0.526 & 0.432\\
    & $f_{sp}^{1-3}$  & $f_{sp}^{1-3}$  & 0.472 & 0.504 & 0.416\\
    & $f_{sp}^{1-5}$  & $f_{sp}^{1-5}$  & 0.478 & 0.512 & 0.418\\
    & w/o $f_{sp}$     & w/o $f_{sp}$     & 0.478 & 0.516 & 0.420\\
    \hline
\end{tabular}
\label{tab:sp}
\vspace{-8pt}
\end{table}

\begin{table}
\caption{Ablation study to verify the effectiveness of Decoder for the proposed RINDNet. SW refers to ``Share Weight'', $1^{st}$ and $2^{nd}$ refer to the first stream and the second stream in Decoder.}
\centering
\small
\renewcommand\tabcolsep{5.0pt}
\renewcommand\arraystretch{1.1}
\begin{tabular}{|cc|cc|ccc|}
    \hline
    \multicolumn{2}{|c|}{RE\&IE-Decoder} &\multicolumn{2}{c|}{NE\&DE-Decoder} &\multicolumn{3}{c|}{Average}\\
    \cline{1-7}
    $1^{st}$ & $2^{nd}$ & $1^{st}$ & $2^{nd}$ & ODS   & OIS   & AP\\
    \hline
    $\surd$ & $\surd$ & $\surd$ & w SW  & \bf{0.486} & \bf{0.526} & \bf{0.432}\\
    $\surd$ & $\times$ & $\surd$ & $\times$  & 0.457 & 0.492 & 0.398\\
    $\surd$ & $\times$ & $\surd$ & w SW  & 0.476 & 0.514 & 0.415\\
    $\surd$ & $\surd$ & $\surd$ & w/o SW     & 0.474 & 0.517 & 0.408\\
    \hline
\end{tabular}
\label{tab:decoder}
\vspace{-13pt}
\end{table}

\begin{figure*}
\centering
\includegraphics[width=0.975\linewidth]{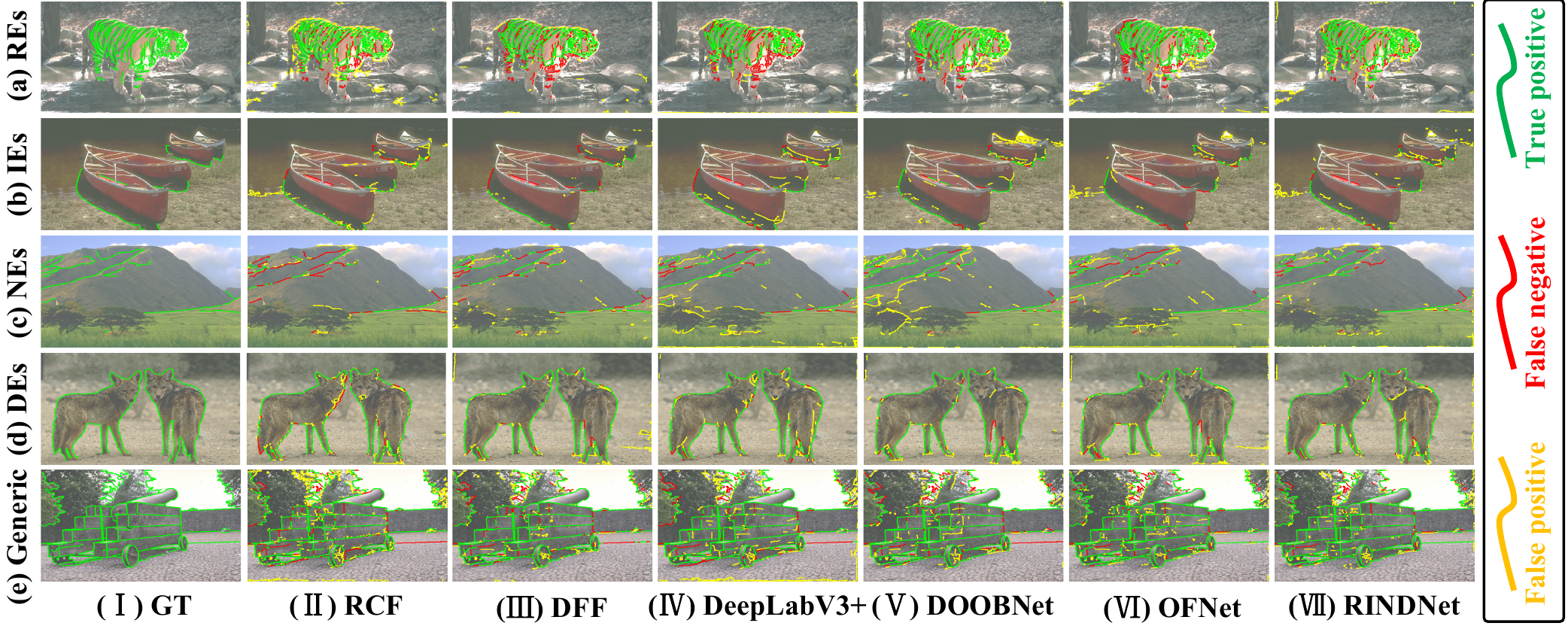}
\caption{Qualitative comparison for (a) Reflectance Edges, (b) Illumination Edges, (c) Normal Edges, (d) Depth Edges and (e) Generic Edges (best viewed in color: ``\textcolor{green2}{\textbf{green}}'' for true positive, ``\textcolor{red}{\textbf{red}}'' for false negative (missing edges), and ``\textcolor{yellow}{\textbf{yellow}}'' for false positive). We provide visualization results of the top 6 scores in this figure, more results can be found in the \textbf{supplementary material}.}
\label{fig:fig4}
\vspace{-13pt}
\end{figure*}

\vspace{-3mm}
\paragraph{Ablation Study.}
We first conduct the ablation study to verify the role of Weight Layer (WL) and Attention Module (AM) in RINDNet. In the experiments, each module is removed separately or together to construct multiple variants for evaluation, as shown in Table~\ref{tab:wlam} (rows 2 -- 4). Intuitively, WL plays a significant role for REs and IEs. Especially for REs, in terms of ODS, OIS and AP, the WL improves the performance significantly from $42.2\%$, $46.8\%$ and $35.7\%$ to $47.8\%$, $52.1\%$ and $41.4\%$, respectively. Besides, with AM, RINDNet achieves noticeable improvements for all types of edges, as shown in row 1 and row 3 of Table~\ref{tab:wlam}. This illustrates the effectiveness of the proposed AM for capturing the distinctions between different edges. Overall, the cooperation of WL and AM allows RINDNet to successfully capture the specific features of each type of edges and thus delivers remarkable performance gain.

Next, we perform an experiment to verify the effectiveness of different-layer features for detecting REs/IEs and NEs/DEs. As shown in Table~\ref{tab:sp} (rows 1 -- 4), the combination based on edge attributes (row 1) performs better than using other choices (rows 3 -- 4). Besides, we also explore the impact of choosing different spatial cues for REs/IEs and NEs/DEs, and report the quantitative results in Table~\ref{tab:sp} (rows 5 -- 7). Similarly, there are average performance drops in different combinations of spatial cues (rows 6 -- 7). Basically, ours (row 5) is the reasonable combination that helps achieve the best performance.

We also design careful ablation experiments (Table \ref{tab:decoder}) to study the effectiveness of each stream in Decoder. The two-stream design combined with share weight (referred to as SW) for NEs/DEs performs better together (row 1) than using either of them separately (rows 3 -- 4). More detailed results are provided in the supplementary material.

\subsection{Experiments on Generic Edges}

To fully examine the performance of RINDNet, we modify and test it on generic edges setting (the ground truth of four types of edges are merged to one ground-truth edge map). The outputs of four decision heads are combined and fed into a final decision head $\psi_{\rm e}$ (one $1\times1$ convolution layer) to predict generic edges: $P=\psi_{\rm e}\big([Y^{r}, Y^{i}, Y^{n}, Y^{d}]\big)$. Moreover, the original ground-truth labels of Attention Module (AM) are unavailable in this setting. Thus we take ground-truth labels of generic edges as the supervisions of AM, so that AM could capture the location of generic edges.

\begin{table}
\caption{Comparison of generic edge detection on \textbf{BSDS-RIND}.
}
\centering
\small
\renewcommand\tabcolsep{14pt}
\renewcommand\arraystretch{0.9}
\begin{tabular}{|l|ccc|}
    \hline
    Method                              & ODS   & OIS   & AP       \\
    \hline
    HED~\cite{xie2015hed}               & 0.786 & 0.805 &\bf{\first{0.834}} \\
    CED~\cite{wang2017ced}              & \bf{\first{0.801}} & \bf{\first{0.814}} & \bf{\second{0.824}} \\
    RCF~\cite{liu2017rcf}               & 0.771 & 0.791 & 0.800 \\
    BDCN~\cite{he2019bdcn}              & 0.789 & 0.803 & 0.757 \\
    DexiNed~\cite{poma2020dexined}      & 0.789	& 0.805	& 0.816 \\
    CASENet~\cite{yu2017casenet}        & 0.792 & 0.806 & 0.786 \\
    DFF~\cite{dff19}                    & 0.794 & 0.806 & 0.767 \\
    DeepLabV3+~\cite{chen2018deeplabv3} & 0.780 & 0.792 & 0.776 \\
    DOOBNet~\cite{wang2018doobnet}      & 0.790 & 0.805 & 0.809 \\
    OFNet~\cite{lu2019ofnet}            & 0.794 & 0.807 & 0.800 \\
    \hline
    RINDNet (Ours)                      &\bf{\second{0.800}} &\bf{\second{0.811}} & 0.815 \\
    \hline
\end{tabular}
\label{tab:generic}
\vspace{-12pt}
\end{table}

We report the quantitative results over generic edges in Table~\ref{tab:generic} and Fig.~\ref{fig:fig3} (e). Note that CED is pre-trained on HED-BSDS. In contrast, our model is trained from scratch and still achieves competitive results. This confirms the integration capability of our model, especially considering that RINDNet is not specially designed for generic edges. Some qualitative results on BSDS-RIND are shown in the last row in Fig. \ref{fig:fig4}.

%------------------------------------------------------------------------
\section{Conclusions}
\label{6}

In this work, we study edge detection on four types of edges including \textit{Reflectance Edges}, \textit{Illumination Edges}, \textit{Normal Edges} and \textit{Depth Edges}. We propose a novel edge detector RINDNet that, for the first time, simultaneously detects all four types of edges. In addition, we contribute the first public benchmark with four types of edges carefully annotated. Experimental results illustrate that RINDNet yields promising results in comparison with state-of-the-art edge detection algorithms.

\vspace{-4mm}
\paragraph{Acknowledgements.}
This work is supported by Fundamental Research Funds for the Central Universities (2019JBZ104) and National Natural Science Foundation of China (61906013, 51827813). The work is partially done while Mengyang Pu was with Stony Brook University.

%-------------------------------------------------------------------------

{\small
\bibliographystyle{ieee_fullname}
\bibliography{egbib}

\begin{thebibliography}{10}\itemsep=-1pt

\bibitem{acuna2019devil}
David Acuna, Amlan Kar, and Sanja Fidler.
\newblock Devil is in the edges: Learning semantic boundaries from noisy
  annotations.
\newblock In {\em IEEE Conf. Comput. Vis. Pattern Recog.}, pages 11075--11083,
  2019.

\bibitem{arbelaez2010bsds}
Pablo Arbelaez, Michael Maire, Charless Fowlkes, and Jitendra Malik.
\newblock Contour detection and hierarchical image segmentation.
\newblock {\em IEEE Trans. Pattern Anal. Mach. Intell.}, 33(5):898--916, 2010.

\bibitem{bertasius2015deepedge}
Gedas Bertasius, Jianbo Shi, and Lorenzo Torresani.
\newblock Deepedge: A multi-scale bifurcated deep network for top-down contour
  detection.
\newblock In {\em IEEE Conf. Comput. Vis. Pattern Recog.}, pages 4380--4389,
  2015.

\bibitem{bertasius2015hfl}
Gedas Bertasius, Jianbo Shi, and Lorenzo Torresani.
\newblock High-for-low and low-for-high: Efficient boundary detection from deep
  object features and its applications to high-level vision.
\newblock In {\em Int. Conf. Comput. Vis.}, pages 504--512, 2015.

\bibitem{canny1986computational}
John~F. Canny.
\newblock A computational approach to edge detection.
\newblock {\em IEEE Trans. Pattern Anal. Mach. Intell.}, 8(6):679--698, 1986.

\bibitem{chen2018deeplabv3}
Liang-Chieh Chen, Yukun Zhu, George Papandreou, Florian Schroff, and Hartwig
  Adam.
\newblock Encoder-decoder with atrous separable convolution for semantic image
  segmentation.
\newblock In {\em Eur. Conf. Comput. Vis.}, pages 801--818, 2018.

\bibitem{cordts2016cityscapes}
Marius Cordts, Mohamed Omran, Sebastian Ramos, Timo Rehfeld, Markus Enzweiler,
  Rodrigo Benenson, Uwe Franke, Stefan Roth, and Bernt Schiele.
\newblock The cityscapes dataset for semantic urban scene understanding.
\newblock In {\em IEEE Conf. Comput. Vis. Pattern Recog.}, pages 3213--3223,
  2016.

\bibitem{DengDSLL009}
Jia Deng, Wei Dong, Richard Socher, Li{-}Jia Li, Kai Li, and Fei{-}Fei Li.
\newblock Imagenet: {A} large-scale hierarchical image database.
\newblock In {\em IEEE Conf. Comput. Vis. Pattern Recog.}, pages 248--255,
  2009.

\bibitem{deng2020dscd}
Ruoxi Deng and Shengjun Liu.
\newblock Deep structural contour detection.
\newblock In {\em ACM Int. Conf. Multimedia}, pages 304--312, 2020.

\bibitem{deng2018lpcb}
Ruoxi Deng, Chunhua Shen, Shengjun Liu, Huibing Wang, and Xinru Liu.
\newblock Learning to predict crisp boundaries.
\newblock In {\em Eur. Conf. Comput. Vis.}, pages 562--578, 2018.

\bibitem{dollar2006supervised}
Piotr Doll{\'{a}}r, Zhuowen Tu, and Serge~J. Belongie.
\newblock Supervised learning of edges and object boundaries.
\newblock In {\em IEEE Conf. Comput. Vis. Pattern Recog.}, volume~2, pages
  1964--1971, 2006.

\bibitem{everingham2010pascal}
Mark Everingham, Luc Van~Gool, Christopher~KI Williams, John Winn, and Andrew
  Zisserman.
\newblock The pascal visual object classes (voc) challenge.
\newblock {\em Int. J. Comput. Vis.}, 88(2):303--338, 2010.

\bibitem{hariharan2011semantic}
Bharath Hariharan, Pablo Arbel{\'a}ez, Lubomir Bourdev, Subhransu Maji, and
  Jitendra Malik.
\newblock Semantic contours from inverse detectors.
\newblock In {\em Int. Conf. Comput. Vis.}, pages 991--998, 2011.

\bibitem{he2019bdcn}
Jianzhong He, Shiliang Zhang, Ming Yang, Yanhu Shan, and Tiejun Huang.
\newblock Bi-directional cascade network for perceptual edge detection.
\newblock In {\em IEEE Conf. Comput. Vis. Pattern Recog.}, pages 3828--3837,
  2019.

\bibitem{he2016deepres}
Kaiming He, Xiangyu Zhang, Shaoqing Ren, and Jian Sun.
\newblock Deep residual learning for image recognition.
\newblock In {\em IEEE Conf. Comput. Vis. Pattern Recog.}, pages 770--778,
  2016.

\bibitem{dff19}
Yuan Hu, Yunpeng Chen, Xiang Li, and Jiashi Feng.
\newblock Dynamic feature fusion for semantic edge detection.
\newblock In {\em IJCAI}, pages 782--788, 2019.

\bibitem{kelm2019rcn}
Andr{\'e}~Peter Kelm, Vijesh~Soorya Rao, and Udo Z{\"o}lzer.
\newblock Object contour and edge detection with refinecontournet.
\newblock In {\em International Conference on Computer Analysis of Images and
  Patterns}, pages 246--258. Springer, 2019.

\bibitem{KimTO15Joint}
Kichang Kim, Akihiko Torii, and Masatoshi Okutomi.
\newblock Joint estimation of depth, reflectance and illumination for depth
  refinement.
\newblock In {\em Int. Conf. Comput. Vis.}, pages 199--207, 2015.

\bibitem{kittler1983accuracy}
Josef Kittler.
\newblock On the accuracy of the sobel edge detector.
\newblock {\em {Image Vis. Comput.}}, 1(1):37--42, 1983.

\bibitem{kokkinos2015pushing}
Iasonas Kokkinos.
\newblock Pushing the boundaries of boundary detection using deep learning.
\newblock {\em Int. Conf. Learn. Represent.}, 2015.

\bibitem{lim2013sketch}
Joseph~J. Lim, C.~Lawrence Zitnick, and Piotr Doll{\'{a}}r.
\newblock Sketch tokens: {A} learned mid-level representation for contour and
  object detection.
\newblock In {\em IEEE Conf. Comput. Vis. Pattern Recog.}, pages 3158--3165,
  2013.

\bibitem{lin2017focal}
Tsung-Yi Lin, Priya Goyal, Ross Girshick, Kaiming He, and Piotr Doll{\'a}r.
\newblock Focal loss for dense object detection.
\newblock In {\em Int. Conf. Comput. Vis.}, pages 2980--2988, 2017.

\bibitem{liu2017rcf}
Yun Liu, Ming-Ming Cheng, Xiaowei Hu, Kai Wang, and Xiang Bai.
\newblock Richer convolutional features for edge detection.
\newblock In {\em IEEE Conf. Comput. Vis. Pattern Recog.}, pages 3000--3009,
  2017.

\bibitem{liu2016rds}
Yu Liu and Michael~S Lew.
\newblock Learning relaxed deep supervision for better edge detection.
\newblock In {\em IEEE Conf. Comput. Vis. Pattern Recog.}, pages 231--240,
  2016.

\bibitem{lu2019ofnet}
Rui Lu, Feng Xue, Menghan Zhou, Anlong Ming, and Yu Zhou.
\newblock Occlusion-shared and feature-separated network for occlusion
  relationship reasoning.
\newblock In {\em Int. Conf. Comput. Vis.}, pages 10343--10352, 2019.

\bibitem{maninis2016cob}
Kevis-Kokitsi Maninis, Jordi Pont-Tuset, Pablo Arbel{\'a}ez, and Luc Van~Gool.
\newblock Convolutional oriented boundaries.
\newblock In {\em Eur. Conf. Comput. Vis.}, pages 580--596. Springer, 2016.

\bibitem{marr1982vision}
David Marr.
\newblock Vision: A computational investigation into the human representation
  and processing of visual information, 1982.

\bibitem{martin2004learning}
David~R. Martin, Charless~C. Fowlkes, and Jitendra Malik.
\newblock Learning to detect natural image boundaries using local brightness,
  color, and texture cues.
\newblock {\em IEEE Trans. Pattern Anal. Mach. Intell.}, 26(5):530--549, 2004.

\bibitem{mely2016multicue}
David~A M{\'e}ly, Junkyung Kim, Mason McGill, Yuliang Guo, and Thomas Serre.
\newblock A systematic comparison between visual cues for boundary detection.
\newblock {\em Vis. Res.}, 120:93--107, 2016.

\bibitem{paszke2017automatic}
Adam Paszke, Sam Gross, Soumith Chintala, Gregory Chanan, Edward Yang, Zachary
  DeVito, Zeming Lin, Alban Desmaison, Luca Antiga, and Adam Lerer.
\newblock Automatic differentiation in pytorch.
\newblock In {\em Adv. Neural Inform. Process. Syst.}, 2017.

\bibitem{poma2020dexined}
Xavier~Soria Poma, Edgar Riba, and Angel Sappa.
\newblock Dense extreme inception network: Towards a robust cnn model for edge
  detection.
\newblock In {\em IEEE Winter Conf. Appl. Comput. Vis.}, pages 1923--1932,
  2020.

\bibitem{qin2018bylabel}
Xuebin Qin, Shida He, Zichen Zhang, Masood Dehghan, and Martin Jagersand.
\newblock Bylabel: A boundary based semi-automatic image annotation tool.
\newblock In {\em IEEE Winter Conf. Appl. Comput. Vis.}, pages 1804--1813,
  2018.

\bibitem{ramamonjisoa2020predicting}
Micha{\"e}l Ramamonjisoa, Yuming Du, and Vincent Lepetit.
\newblock Predicting sharp and accurate occlusion boundaries in monocular depth
  estimation using displacement fields.
\newblock In {\em IEEE Conf. Comput. Vis. Pattern Recog.}, pages 14648--14657,
  2020.

\bibitem{ren2006figure}
Xiaofeng Ren, Charless~C Fowlkes, and Jitendra Malik.
\newblock Figure/ground assignment in natural images.
\newblock In {\em Eur. Conf. Comput. Vis.}, pages 614--627, 2006.

\bibitem{Seitz04wi}
Steven Seitz.
\newblock Edge detection.
\newblock
  \url{https://courses.cs.washington.edu/courses/cse455/04wi/lectures/edge.pdf}.

\bibitem{shen2015deepcontour}
Wei Shen, Xinggang Wang, Yan Wang, Xiang Bai, and Zhijiang Zhang.
\newblock Deepcontour: A deep convolutional feature learned by positive-sharing
  loss for contour detection.
\newblock In {\em IEEE Conf. Comput. Vis. Pattern Recog.}, pages 3982--3991,
  2015.

\bibitem{silberman2012indoor}
Nathan Silberman, Derek Hoiem, Pushmeet Kohli, and Rob Fergus.
\newblock Indoor segmentation and support inference from rgbd images.
\newblock In {\em Eur. Conf. Comput. Vis.}, pages 746--760, 2012.

\bibitem{wang2018doobnet}
Guoxia Wang, Xiaochuan Wang, Frederick~WB Li, and Xiaohui Liang.
\newblock Doobnet: Deep object occlusion boundary detection from an image.
\newblock In {\em ACCV}, pages 686--702. Springer, 2018.

\bibitem{wang2016doc}
Peng Wang and Alan Yuille.
\newblock Doc: Deep occlusion estimation from a single image.
\newblock In {\em Eur. Conf. Comput. Vis.}, pages 545--561, 2016.

\bibitem{wang2015designing}
Xiaolong Wang, David~F. Fouhey, and Abhinav Gupta.
\newblock Designing deep networks for surface normal estimation.
\newblock In {\em IEEE Conf. Comput. Vis. Pattern Recog.}, pages 539--547,
  2015.

\bibitem{wang2017ced}
Yupei Wang, Xin Zhao, and Kaiqi Huang.
\newblock Deep crisp boundaries.
\newblock In {\em IEEE Conf. Comput. Vis. Pattern Recog.}, pages 3892--3900,
  2017.

\bibitem{winnemoller2011xdog}
Holger Winnem{\"{o}}ller, Jan~Eric Kyprianidis, and Sven~C. Olsen.
\newblock Xdog: An extended difference-of-gaussians compendium including
  advanced image stylization.
\newblock {\em Comput. Graph.}, 36(6):740--753, 2012.

\bibitem{wu2012strong}
Qi Wu, Wende Zhang, and B.~V. K.~Vijaya Kumar.
\newblock Strong shadow removal via patch-based shadow edge detection.
\newblock In {\em International Conference on Robotics and Automation}, pages
  2177--2182, 2012.

\bibitem{xie2015hed}
Saining Xie and Zhuowen Tu.
\newblock Holistically-nested edge detection.
\newblock In {\em Int. Conf. Comput. Vis.}, pages 1395--1403, 2015.

\bibitem{xu2017AMHNet}
Dan Xu, Wanli Ouyang, Xavier Alameda-Pineda, Elisa Ricci, Xiaogang Wang, and
  Nicu Sebe.
\newblock Learning deep structured multi-scale features using attention-gated
  crfs for contour prediction.
\newblock In {\em Adv. Neural Inform. Process. Syst.}, pages 3961--3970, 2017.

\bibitem{YangZYPML20}
Fan Yang, Lei Zhang, Sijia Yu, Danil Prokhorov, Xue Mei, and Haibin Ling.
\newblock Feature pyramid and hierarchical boosting network for pavement crack
  detection.
\newblock {\em IEEE Trans. Intell. Transp. Syst.}, 21(4):1525--1535, 2019.

\bibitem{yang2016object}
Jimei Yang, Brian Price, Scott Cohen, Honglak Lee, and Ming-Hsuan Yang.
\newblock Object contour detection with a fully convolutional encoder-decoder
  network.
\newblock In {\em IEEE Conf. Comput. Vis. Pattern Recog.}, pages 193--202,
  2016.

\bibitem{yu2017casenet}
Zhiding Yu, Chen Feng, Ming{-}Yu Liu, and Srikumar Ramalingam.
\newblock Casenet: Deep category-aware semantic edge detection.
\newblock In {\em IEEE Conf. Comput. Vis. Pattern Recog.}, pages 5964--5973,
  2017.

\bibitem{yu2018simultaneous}
Zhiding Yu, Weiyang Liu, Yang Zou, Chen Feng, Srikumar Ramalingam, B.~V.
  K.~Vijaya Kumar, and Jan Kautz.
\newblock Simultaneous edge alignment and learning.
\newblock In {\em Eur. Conf. Comput. Vis.}, pages 388--404, 2018.

\bibitem{zhen2020joint}
Mingmin Zhen, Jinglu Wang, Lei Zhou, Shiwei Li, Tianwei Shen, Jiaxiang Shang,
  Tian Fang, and Long Quan.
\newblock Joint semantic segmentation and boundary detection using iterative
  pyramid contexts.
\newblock In {\em IEEE Conf. Comput. Vis. Pattern Recog.}, pages 13666--13675,
  2020.

\end{thebibliography}
}

\end{document}